\documentclass[journal,twoside]{IEEEtran}
\usepackage{cite}
\usepackage{amsmath,amssymb,amsfonts}
\usepackage{algorithmic}
\usepackage{graphicx}
\usepackage{textcomp}
\usepackage{bm}
\usepackage{array}
\usepackage{stfloats}
\usepackage{url}
\usepackage{verbatim}
\usepackage[hidelinks]{hyperref}
\usepackage{booktabs}
\usepackage[ruled,linesnumbered]{algorithm2e}
\usepackage{multirow}
\usepackage{makecell}
\usepackage{adjustbox}

\def\BibTeX{{\rm B\kern-.05em{\sc i\kern-.025em b}\kern-.08em
    T\kern-.1667em\lower.7ex\hbox{E}\kern-.125emX}}

\begin{document}

\title{Efficient Extreme Operating Condition Search for Online Relay Setting Calculation in Renewable Power Systems Based on Graph Neural Network}
\author{
    Yan~Li, %
    Zengli~Yang, %
    Youhuai~Wang, %
    Jing~Wang, %
    Xiaoyu~Han, %
    Jingyu~Wang,~\IEEEmembership{Member,~IEEE}, %
    Yinhong~Li,~\IEEEmembership{Senior~Member,~IEEE}, %
    Dongyuan~Shi,~\IEEEmembership{Senior~Member,~IEEE}%
    \thanks{
      The work was supported by the National Natural Science Foundation of China (Grant No. 52207107). %
      \emph{(Corresponding author: Jingyu Wang)}
    }
     \thanks{
       Yan Li, Xiaoyu Han, Jingyu Wang, Yinhong Li, and Dongyuan Shi are with the State Key Laboratory of Advanced Electromagnetic Technology and the School of Electrical and Electronic Engineering, Huazhong University of Science and Technology, Wuhan, China (e-mails: \href{xmlee@hust.edu.cn}{xmlee@hust.edu.cn}, \href{xiaoyu_han@hust.edu.cn}{xiaoyu\_han@hust.edu.cn}, \href{jywang@hust.edu.cn}{jywang@hust.edu.cn}, \href{liyinhong@hust.edu.cn}{liyinhong@hust.edu.cn}, \href{dongyuanshi@hust.edu.cn}{dongyuanshi@hust.edu.cn}). 
      
       Zengli Yang, Youhuai Wang, and Jing Wang are with the State Grid Hubei Electric Power Co., Ltd., Wuhan, 430074, Hubei, China (e-mails: \href{yangzl_bh@163.com}{yangzl\_bh@163.com}, \href{13871327816@139.com}{13871327816@139.com}, \href{58099687@qq.com}{58099687@qq.com}).
     }
}
\maketitle

\begin{abstract}
The Extreme Operating Conditions Search (EOCS) problem is one of the key problems in relay setting calculation, which is used to ensure that the setting values of protection relays can adapt to the changing operating conditions of power systems over a period of time after deployment. The high penetration of renewable energy and the wide application of inverter-based resources make the operating conditions of renewable power systems more volatile, which urges the adoption of the online relay setting calculation strategy. However, the computation speed of existing EOCS methods based on local enumeration, heuristic algorithms, and mathematical programming cannot meet the efficiency requirement of online relay setting calculation. To reduce the time overhead, this paper, for the first time, proposes an efficient deep learning-based EOCS method suitable for online relay setting calculation. First, the power system information is formulated as four layers, i.e., a component parameter layer, a topological connection layer, an electrical distance layer, and a graph distance layer, which are fed into a parallel graph neural network (PGNN) model for feature extraction. Then, the four feature layers corresponding to each node are spliced and stretched, and then fed into the decision network to predict the extreme operating condition of the system. Finally, the proposed PGNN method is validated on the modified IEEE 39-bus and 118-bus test systems, where some of the synchronous generators are replaced by renewable generation units. The nonlinear fault characteristics of renewables are fully considered when computing fault currents. The experiment results show that the proposed PGNN method achieves higher accuracy than the existing methods in solving the EOCS problem. Meanwhile, it also provides greater improvements in online computation time.
\end{abstract}


\section{Introduction}
\label{sec:introduction}
\IEEEPARstart{R}{elay} protection serves as a critical safeguard in power systems, enabling the detection and isolation of faulty equipment to ensure the safety and stability of electricity transfer. Upon fault occurrence, protection relays activate circuit breakers to contain fault propagation and mitigate widespread power interruptions \cite{1645145}. In current engineering practice, the relay setting calculation process begins with the definition of a comprehensive set of anticipated operating conditions, which is typically derived from the routine operating scenarios of the power system. Specifically, the construction of the anticipated operating condition set involves a traversal process that starts from the baseline maximum/minimum operating conditions of the power system and enumerates different combinations of the on/off states of equipment in the range of certain levels adjacent to the target protection device whose setting values are to be determined. In addition to the traversal process, some special operating conditions, manually characterized from historical operating experience, can also be added to the set. Very special operating scenarios, such as extreme disasters or rare events, are usually not considered when constructing the envisioned operating condition set. Critical operating conditions are then identified from the anticipated set that induce extreme values of fault-related parameters, including short-circuit currents, measured impedance, branching/auxiliary increasing coefficients, at the intended protection location. Under these critical conditions, careful coordination is performed between the target protection device and its upstream/downstream counterparts, culminating in an appropriate set of relay settings \cite{7081386}.

In the above relay setting calculation process, the identified critical operating conditions that cause certain electrical quantities to reach extreme values are defined as the Extreme Operating Conditions (EOCs) of the protection. Mathematically, the EOC Search (EOCS) problem is a combinatorial optimization problem \cite{7347480}. Currently, relay settings are predominantly computed offline in advance and kept invariant after commissioning. This offline paradigm does not impose stringent requirements on the computational efficiency of the relay setting calculation. As a result, methods typically used to solve combinatorial optimization problems, such as local enumeration, heuristic algorithms \cite{4682639}, and mathematical programming \cite{10841900}, are often employed to solve the EOCS problem in practical relay setting calculation applications \cite{1256359}. The local enumeration method involves the evaluation of EOCs within limited protection zones or under specific fault scenarios. In \cite{10266989}, an enumeration-based current differential protection scheme accelerates the setting calculation by restricting the search space according to global optimum distribution patterns in non-convex constraints. In \cite{SHEN2024110503}, a streamlined EOC derivation approach is achieved through selective fault case analysis for different characteristic quantities. In \cite{9172105}, transmission line protection integrity is improved by tabulating operating states and fault conditions to quantify relay margins and optimize settings. In \cite{9696364}, a gray wolf algorithm-based overcurrent protection coordination framework for wind farms using meta-heuristic optimization is proposed. In \cite{7355377}, a distribution system protection scheme uses Rudin-Osher-Fatemi modeling for equivalent circuit parameter estimation. In \cite{7359148}, the coordination of directional overcurrent relays in distributed systems is formulated as a nonlinear programming problem solved by Ipopt and Baron solvers. However, the scalability of heuristic algorithms and mathematical programming methods remains a challenge, as their convergence time increases with the increased dimension of the problem to be solved \cite{9863874}.

Renewable power systems have highly flexible and volatile operating dynamics with frequent occurrence of atypical operating scenarios, rendering the conventional offline relay setting calculation strategy inadequate due to their inability to adapt to rapid operating state changes \cite{8666728}. For renewable power systems, offline relay setting calculation faces two critical limitations: reliance on predefined baseline operating conditions and ignorance of extreme operating scenarios. On the one hand, the offline relay setting calculation uses the baseline operating conditions of the system as a starting point for the traversal construction of the anticipated operating condition set. However, the highly random nature of sources and loads in renewable power systems makes it difficult to clearly define baseline operating conditions. On the other hand, extreme operating scenarios are typically not considered in the offline relay setting calculation to avoid sacrificing the performance of the settings under the majority of common operating conditions. Nonetheless, the high penetration of renewable generation units and power electronic devices makes renewable power systems more susceptible to extreme weather and operational disturbances, leading to more frequent atypical operating scenarios. Neglecting special operating conditions in the calculation process will significantly degrade the adaptability of the obtained settings when extreme operating scenarios occur. To address these challenges, renewable power systems require implementation of the online relay setting calculation paradigm, which continuously monitors the operating condition changes of power systems and triggers relay setting calculations when the setting values cannot adapt to the current operating condition. Compared to offline relay setting calculation, online calculation enables dynamic adjustments to protection parameters through more frequent updates based on evolving system states, significantly enhancing adaptability to operational uncertainties of renewable power systems \cite{10373986}.

Online relay setting calculation imposes a strong efficiency requirement for solving the EOCS problem. The high dimensionality of the combinatorial search domain hinders the application of existing local enumeration and combinatorial optimization methods to timely find the EOCs \cite{8025725}. Worse still, existing approaches rely on a large number of short-circuit calculations under various operating conditions to guide the search for EOCs. Due to the nonlinear current-source fault characteristics of renewable generation units, short-circuit calculations have to be performed in an iterative manner in renewable power systems \cite{FAN2025110418}. The complication of the short-circuit calculation process further reduces the viability of traditional EOCS methods for online application. A transformative methodology that decouples relay setting calculations from high-dimensional combinatorial optimization and large-scale short-circuit calculations is urgently needed.

Over the past decade, artificial intelligence technologies, in particular deep learning (DL), have made remarkable progress and found extensive applications in various domains of power systems \cite{DONON2020106547, 10070812, 10669036, 10528868}. DL divides the decision making process into an offline training phase and an online application phase. During the offline training phase, DL frameworks learn a comprehensive mapping between the input features and the output labels, thereby significantly reducing the computational overhead during the online application. By using DL techniques to learn the complex mapping between the input features of renewable power systems, e.g., the topology and the protection installation location, and the corresponding EOCs, the EOCS problem can be expected to be solved with very short online time costs \cite{9583605}.

This paper presents a DL-based EOCS method based on Parallel Graph Neural Network (PGNN). The proposed PGNN model first extracts the features of renewable power systems, including components, topology, electrical distance and graph distance, and then directly predicts the EOC based on the input feature. The main contributions of this paper include:
\begin{enumerate}
    \item{This paper, for the first time, employs DL techniques to solve the EOCS problem. Unlike conventional approaches requiring exhaustive evaluations of operating conditions and intensive short-circuit calculations, the proposed DL-based method accelerates the online prediction of EOCs via offline mapping learning, making it more suitable for online relay setting calculation.}
    \item{Four feature encoding modules are used to characterize the input features of renewable power systems. A PGNN architecture is designed to extract latent information from the input features, where each type of feature is processed by independent graph neural networks (GNNs) to achieve a finer granularity of learning. The processed features are then integrated and fed into a decision network to finally output the EOC.}
    \item{The proposed method fully considers the specificity of renewable power systems in the EOCS problem. Different feature encodings are implemented for synchronous generators and renewable generation units to enable the discrimination of both types of generation during learning. Besides, the nonlinear voltage-controlled current-source fault characteristics of renewable generation units are considered in the training sample generation process, ensuring accurate calculation of short-circuit currents.}
\end{enumerate}

The rest of the paper is organized as follows. Section \ref{sec:PRELIMINARY TECHNOLOGY} introduces the preliminary knowledge of GNN and the principle of short-circuit fault analysis for renewable power systems. Section \ref{sec:PGNN} describes the proposed feature encoding method and the PGNN framework. Section \ref{sec:EXPERIMENTS} validates and compares the performance of PGNN in solving the EOCS problem. Section \ref{sec:CONCLUSION} concludes the paper.

\section{Preliminary Knowledge}
\label{sec:PRELIMINARY TECHNOLOGY}
This section introduces the short-circuit calculation method for renewable power systems and the basic principle of graph neural networks.

\subsection{Short-Circuit Calculation for Renewable Power Systems}
\label{sec:short-circuit calculation}
The short-circuit calculation method based on the node conductance matrix is one of the most commonly used methods. It is based on the superposition theorem, where the fault component $\Delta \mathbf{Y}^{(f)}$ is superimposed on the pre-fault three-sequence conductivity matrices $\mathbf{Y}_{120}^{(0)}$. The three-sequence current injection $\bm{I}_{120}$ is assumed to be unchanged at the fault moment. Thus, the three-sequence voltage $\bm{V}_{120}$ can be solved.

For renewable power systems, nodes can be categorized into three types based on their fault characteristics: synchronous generator nodes, renewable generation unit nodes, and other nodes. In the short-circuit calculation, synchronous generators are modeled as constant voltage sources, while renewable generation units are modeled as nonlinear voltage controlled current sources.
The short circuit current calculation model of renewable generation units comprises two distinct models as follows \cite{FAN2025110418}, and the voltage-current mapping relationships during a symmetrical short circuit of these two models are governed by different piecewise equations expressed as Fig. \ref{fig:FIPS} and Fig. \ref{fig:PIPS}.


  

\begin{enumerate}
    \item{Full-Power Inverter Power Supply (FIPS) Model including photovoltaic power plants and permanent magnet synchronous generators.
    As shown in Fig. \ref{fig:FIPS}, $U_1$ is the positive-sequence voltage at the machine terminals, $m$ is the ratio of the output power to the rated capacity before the fault, and $I_{lim}$ is the upper limit of the output current.}
    \begin{figure}
    \centering
    \includegraphics[width=3.2in]{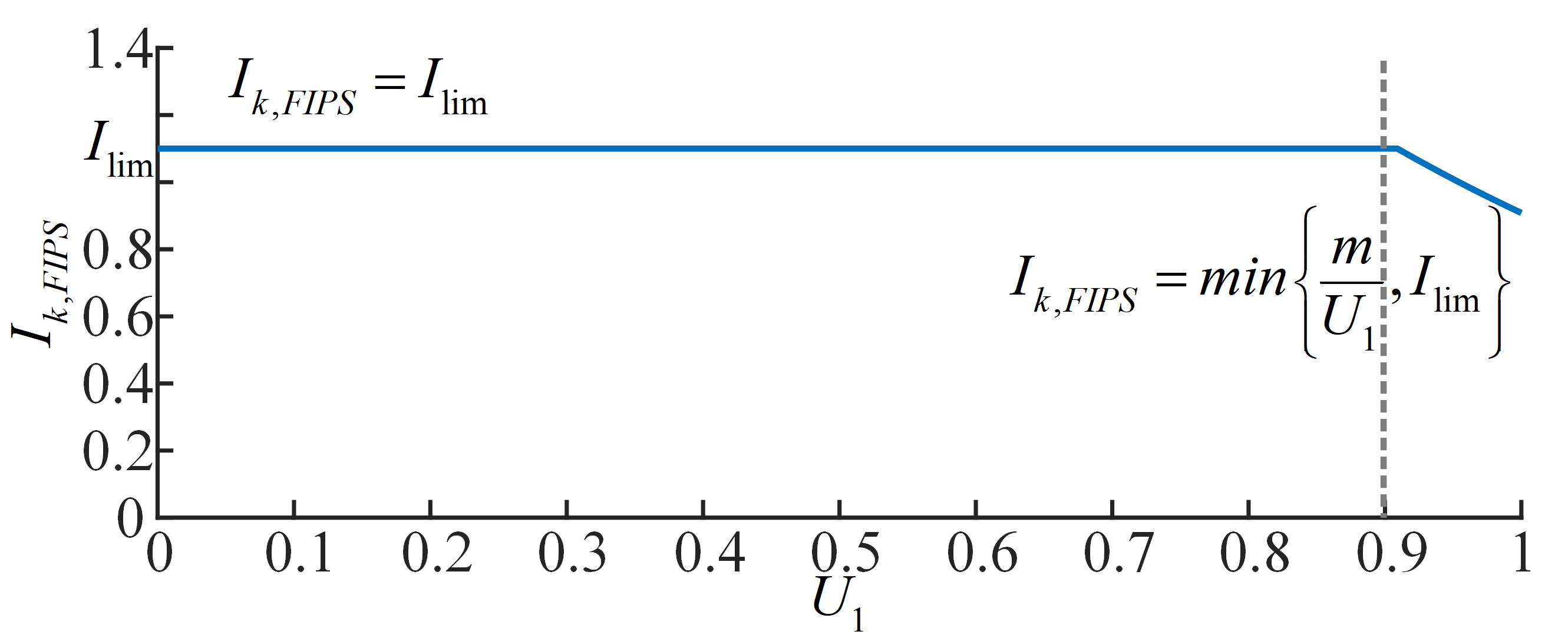}
    \caption{Voltage–current characteristics of PIPS.}
    \label{fig:FIPS}
    \end{figure}
    
    \item{Partial-Power Inverter Power Supply (PIPS) Model represented by double fed induction generator wind turbines.
    As shown in Fig. \ref{fig:PIPS}, $U$ is the standard voltage, $P_0$ is the active power before the fault, and $I_{dcb}$ and $I_{qcb}$ are equivalent impedance coefficients, which are determined by the operation mode of PIPS.}
    \begin{figure}
    \centering
    \includegraphics[width=3.2in]{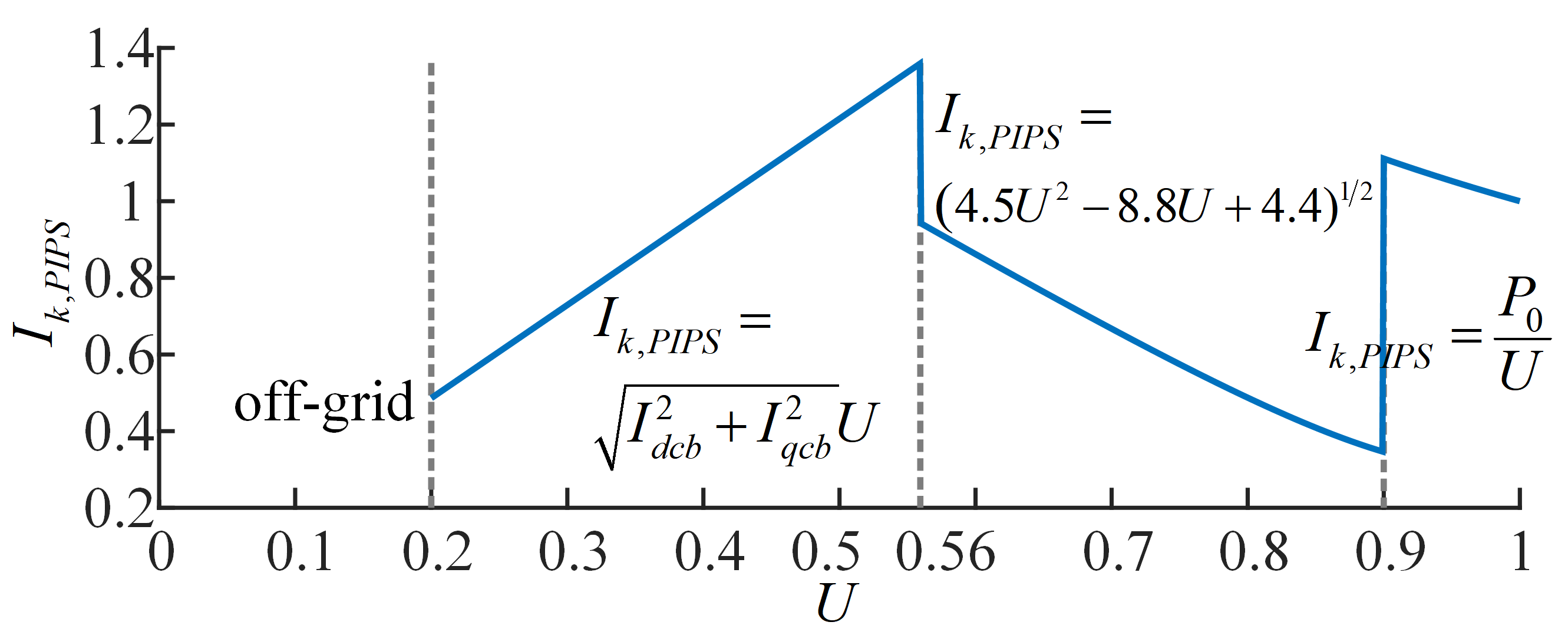}
    \caption{Voltage–current characteristics of PIPS.}
    \label{fig:PIPS}
    \end{figure}
\end{enumerate}



The short-circuit calculation of power systems with renewable generation units is no longer about solving a set of linear equations. At the moment of the fault, both the output current and the terminal voltage of the renewable generation units will change. Therefore, it is necessary to iteratively solve for the magnitude of the short-circuit current \cite{9682675}.

Additionally, in short-circuit calculations, only the injected currents from renewable generation units within an $n$-level range near the fault point need to be considered, with those outside the $n$-level range set to zero. This paper takes \( n=3 \). Therefore, for the aforementioned three types of nodes, the injected currents can be computed as
\begin{equation}
I_i = \left\{
\begin{array}{ll}
\frac{1}{x_{d,i}^{''}} & ,i \in \mathcal{G} \\
I_{k,FIPS}~\text{or}~I_{k,PIPS} & ,i \in \mathcal{R} \text{ and } i \in \mathcal{N}_n(f) \\
0 & ,\text{others}
\end{array}
\right.
\label{eq:If}
\end{equation}
where $\mathcal{G}$ is the set of generator nodes, $\mathcal{R}$ is the set of renewable generation unit nodes, $\mathcal{N}_n(f)$ is the set of nodes within $n$-level of the faulty node, $x_{d,i}^{''}$ is the subtransient reactance of the synchronous generator.

\subsection{Graph Neural Network}
\label{subsec:GNN}
\begin{figure}
\centering
\includegraphics[width=3.2in]{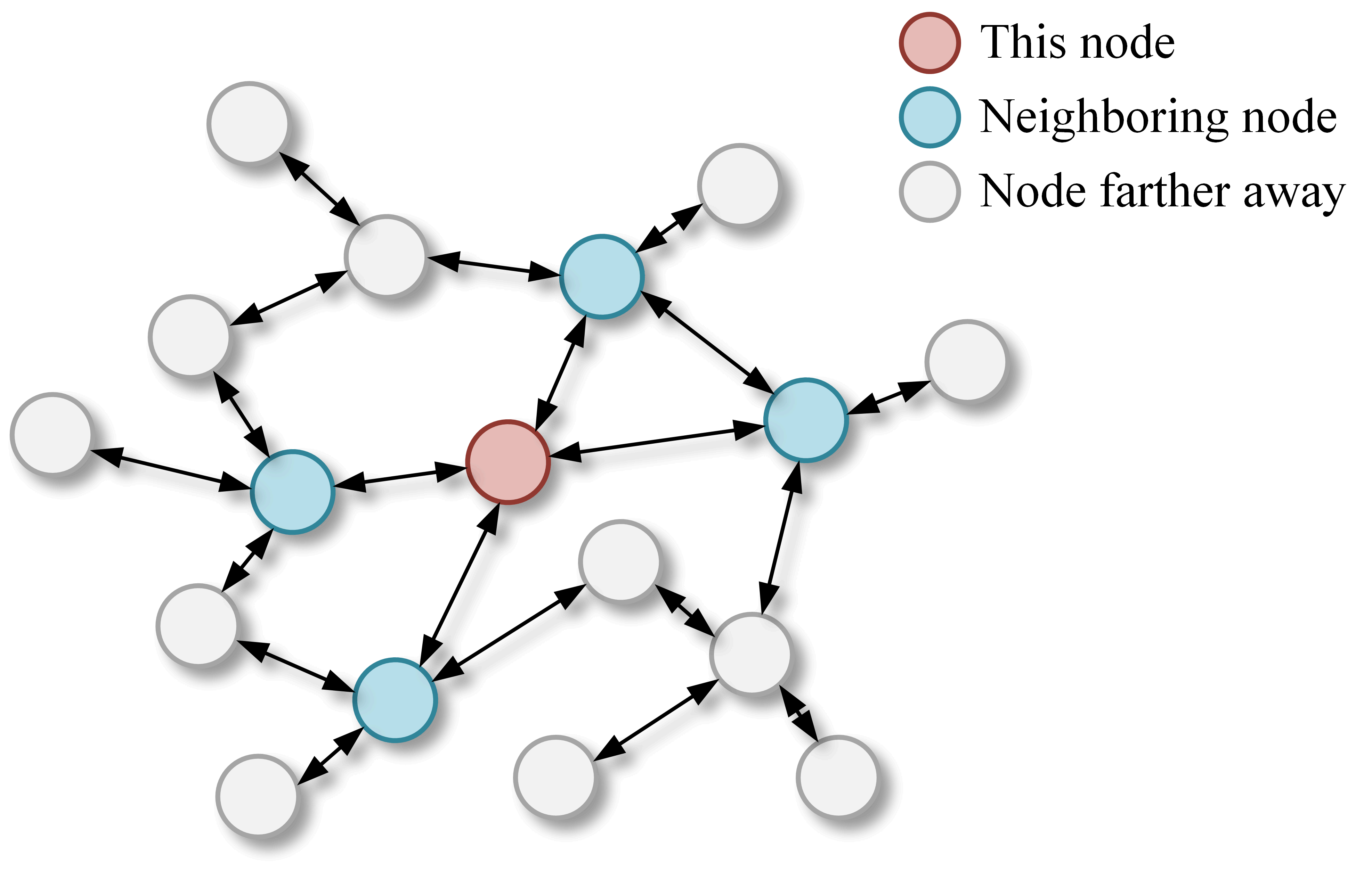}
\caption{Schematic diagram of GNN principle.}
\label{fig:GNN}
\end{figure}

Graph Neural Networks (GNNs) \cite{2009TheGNN} are a type of neural network designed to process graph-structured data that includes nodes and edges, as shown in Fig. \ref{fig:GNN}. Unlike Convolutional Neural Networks (CNNs), which aggregate information from adjacent pixels, GNNs operate based on a message-passing mechanism, aggregating information according to node connections. In a GNN layer, each node updates its features based on its own information and the information from its neighboring nodes, allowing messages to propagate through the network. By stacking multiple layers together, the features of each nodes will ultimately encompass information from all nodes within a certain number of adjacent levels.

GraphSAGE is a basic GNN that leverages node feature information to efficiently generate node embeddings for previously unseen data \cite{graphsage}. Updated features of nodes can be computed as
\begin{equation}
\begin{aligned}
\bm{h}_v^{(l+1)} 
= \sigma\left( \mathbf{W}^{(l)} \cdot \mathrm{AGG}^{(l)}\left( \{ \bm{h}_u^{(l)}, \forall u \in \mathcal{N}(v) \} \right) \right)
\end{aligned}
\label{eq:GraphSage}
\end{equation}
\begin{equation}
\mathrm{AGG}\left ( \bm{h}_u^{(l)},\forall u \in \mathcal{N}(v) \right )  = \left\{
\begin{array}{ll}
\underset{u \in \mathcal{N}(v)}{\sum}\bm{h}_u^{(l)} & ,\text{summation} \\
\frac{1}{\left | \mathcal{N}(v) \right | }\underset{u \in \mathcal{N}(v)}{\sum}\bm{h}_u^{(l)} & ,\text{averaging}   \\
\underset{u \in \mathcal{N}(v)}{\max}\bm{h}_u^{(l)} & ,\text{max-pooling} \\
~~...
\end{array}
\right.
\label{eq:AGG}
\end{equation}
where $\bm{h}_v^{(l)}$ is the node feature vector of node $v$ in layer $l$, $\mathbf{W}^{(l)}$ is the weight matrix to be learned, and $\mathcal{N}(v)$ is the set of neighboring nodes of $v$. $\mathrm{AGG}(\cdot)$ is the aggregation function, which is used to aggregate the features of the connected nodes, which can be implemented by summation, averaging, max-pooling, or other operations, as shown in \eqref{eq:AGG}.

Power systems can be modeled as graphs and described in graph-structured data. Graph Neural Networks (GNNs) are a class of neural networks that specialize in processing graph-structured data. GNNs perform feature aggregation based on the connectivity of nodes and are suitable for processing non-Euclidean data such as power systems. In contrast, traditional machine learning models, represented by Convolutional Neural Networks (CNNs), typically aggregate features based on the spatial positions of pixels and are suitable for processing Euclidean data \cite{9451544}. However, the node numbers of power systems are defined artificially and do not possess spatial features, making them non-Euclidean data. Therefore, GNNs can better extract the features of power systems \cite{9520300}.

\section{Proposed PGNN Framework for Solving EOCS Problems}
\label{sec:PGNN}
This section establishes a mathematical framework for solving an EOCS problem to guarantee the selectivity of instantaneous current protections. A combinatorial optimization model addressing this problem is developed first through the formulation of an objective function with operational constraints. To solve the EOCS problem, two key innovations are developed: (1) a parallel feature coding method for multidimensional data representation, and (2) a parallel graph neural network architecture specifically designed for hierarchical feature extraction.
The proposed framework encodes power system information, including component parameters, topological relationships, electrical distance metrics, and graph connectivity patterns into four matrices. Through parallel feature extraction via graph neural networks, the PGNN learns the latent representations of power system characteristics. These learned features are then processed through a specialized decision network that computes optimal EOCs.

\subsection{Modelling the EOCS Problem}
EOCS specifically refers to searching for the operating condition that causes the protection selectivity or sensitivity reach a critical point for a given protection under a certain initial operating condition of the system. It is used to calculate preparatory quantities required for the setting calculations. This paper takes the selectivity EOCS of instantaneous overcurrent protection as an example for research.

Instantaneous overcurrent protection is a simple and commonly used transmission line protection that detects the occurrence of short-circuit faults by detecting whether the current flowing through the line is greater than the setting value. The setting principle is that the setting value of the protection device at the first end of the line is slightly higher than the maximum short-circuit current at the end of the line to maintain the protection selectivity. Therefore, EOCs are the operating conditions that make the short-circuit current at the end of the lines reach its global maximum over all possible operating conditions. The setting value $I_{set}$ and the tripping criterion are as follows
\begin{equation}
I_{set}=K_{f}I_{f\cdot max \cdot end}
\label{eq:Iset}
\end{equation}
\begin{equation}
I>I_{set}
\label{eq:I>Iset}
\end{equation}
Where $I_{set}$ is the protection setting value, $K_{f}$ is the reliability coefficient, usually taken as 1.2 to 1.3, $I_{f\cdot max \cdot end}$ is the maximum short-circuit current measured by the protection when the short-circuit fault occurs at the end of the line, and is usually considered to be the three-phase short-circuit current at the end of the line under the maximum operating conditions of the system. Therefore, the setting calculation of instantaneous current protection must first obtain the preparatory quantities $I_{f\cdot max \cdot end}$. Meanwhile, it is generally assumed that EOCs are carried out in the range of $N-k$, i.e., the outage of at most $k$ lines is considered. Such an assumption is also reasonable because the probability of multiple faults occurring simultaneously in the system is extremely low and the short-circuit current does not increase indefinitely with the number of failed lines. Let $I_f$ be the fault current flowing through the line at the occurrence of fault $f$. A mathematical model of the EOCS problem can be developed as follows
\begin{equation}
\begin{aligned}
I_{f\cdot max \cdot end}=\max I_f=\max \mathrm{F}_{scc} \left( \bm{\tau}|\bm{\tau}_0,p_{l},f  \right)
\end{aligned}
\label{eq:maxif}
\end{equation}

\begin{equation}
\mathrm{s.t.}\left\{
\begin{array}{ll}
\bm{\tau}_0 \in \{0, 1\}^m \\ \bm{\tau}\in \{0, 1\}^m\\
\|\bm{\tau}_0\|_0 - \|\bm{\tau}\|_0 \leq k\\
p_l \in \{1, 2, \dots, m\}\\
f = f^{(3)}
\end{array}
\right.
\label{eq:s.t.}
\end{equation}
where $\mathrm{F}_{scc}$ is the function for solving short-circuit current, $\bm{\tau}_0$ is the initial operating condition vector of the system, and $\bm{\tau}$ represents the EOC vector that needs to be solved. In $\bm{\tau}_0$ and $\bm{\tau}$, element values equal to 1 indicates the corresponding lines are on operation, while element values equal to 0 indicates the corresponding lines are out-of-operation. The problem is to calculate the EOC of the protection at the start of the transmission line $p_l$. $f=f^{(3)}$ indicates that only three-phase short-circuit faults at the end of $p_l$ are considered.

\begin{figure}[!t]
\centering
\includegraphics[width=3in]{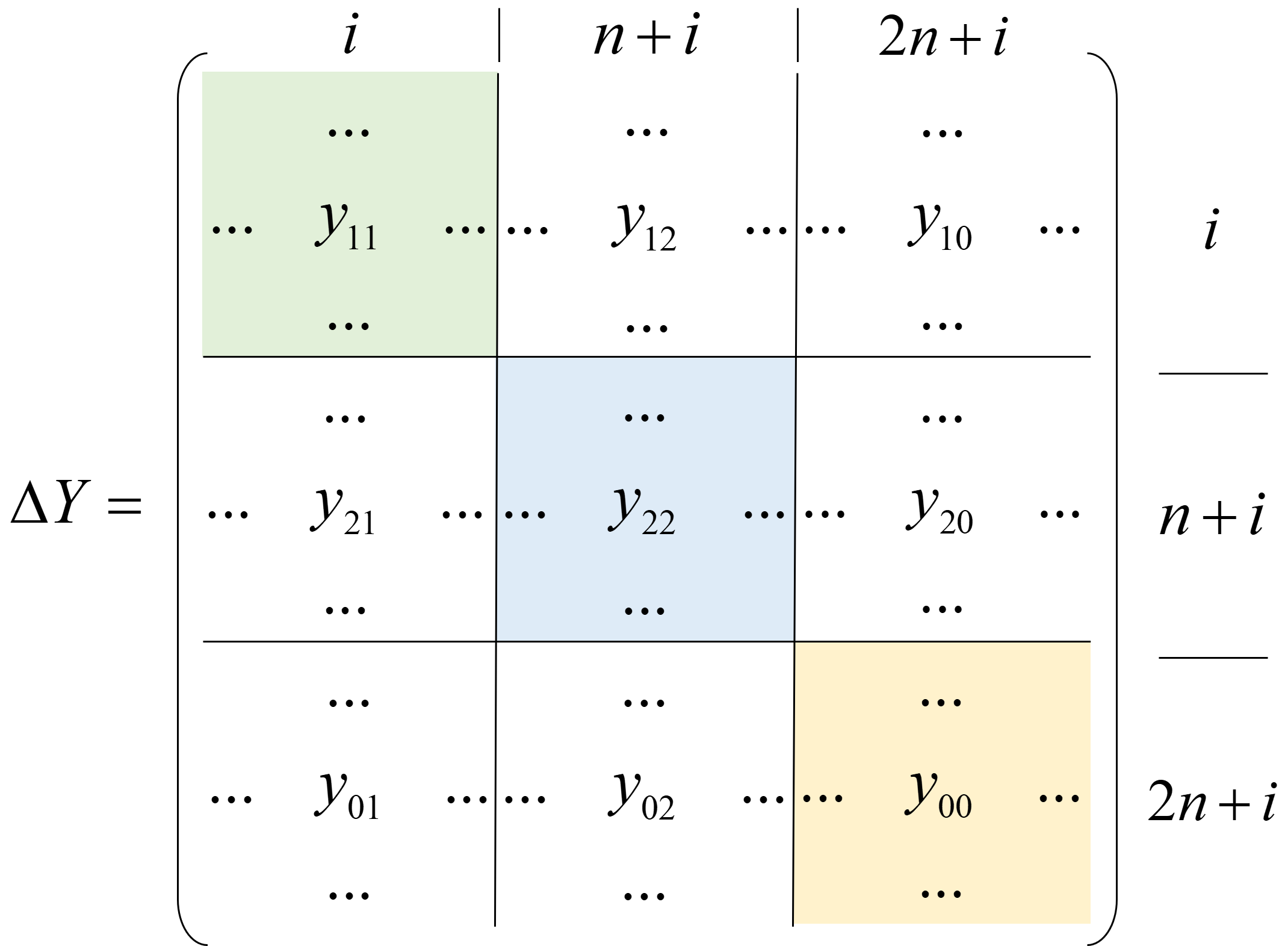}
\caption{The modified components of the admittance matrix of three-phase fault.}
\label{fig:Y-y}
\end{figure}

The form of $\Delta \mathbf{Y}^{(f)}$ of the three-phase short-circuit faults is simple and is determined by the following $\Delta \mathbf{y}$
\begin{equation}
\Delta \mathbf{y}=\begin{bmatrix}
 y_{11} &  y_{12} &  y_{10}\\
 y_{21} &  y_{22} &  y_{20}\\
 y_{01} &  y_{02} &  y_{00}\\
\end{bmatrix}_{3\times 3}=\begin{bmatrix}
 M &   & \\
 &  M &  \\
 &   &  M\\
\end{bmatrix}_{3\times 3}
    \label{eq:dy}
\end{equation}
From a strict physical derivation, $M$ should be infinite, which in practical implementation can be introduced by a large number such as $1e6$. Each of the nine values in $\Delta \mathbf{y}$ corresponds to the value in each of the nine regions in $\Delta \mathbf{Y}^{(f)}$, as shown in Fig. \ref{fig:Y-y}, where $i$ is the faulty node.

\subsection{Parallel Feature Coding Method}
In this section, a parallel feature coding method is proposed to encode information of power system from four categories including component parameters, topology, electrical distance, and graph distance. The four categories of information are firstly constructed in the form of a square matrix of equal dimensions respectively, and then each row of the square matrices is input into the GNNs as a feature vector of the corresponding node. The dimension of the square matrices is the same as the number of buses in the power system.
\begin{enumerate}
    \item{\textbf{Component Parameter Matrix $\mathbf{P}$:} The Component Parameter Matrix contains information about the types and parameters of components in the grid. The grid contains five types of components: synchronous generators, renewable generation units, transmission lines, transformers, and buses. The diagonal elements of the matrix indicate the type of node and the non-diagonal elements indicate the impedance of the branch connecting the two nodes, as follows}
    \begin{equation}
    \mathbf{P}_{ii}=\left\{
    \begin{array}{ll}
    1 &,i \in \mathcal{G} \\
    0.75 &,i \in \mathcal{R} \\
    0.5 &,i \in \mathcal{T}_H \\
    0.25 &,i \in \mathcal{T}_L  \\
    0 &,\text{others} \\
    \end{array}
    \right.
    ,
    \mathbf{P}_{ij}=\left\{
    \begin{array}{ll}
    z_{b(i,j)} &,j \in \mathcal{N}(i) \\
    0 &,j \notin \mathcal{N}(i) \\
    \end{array}
    \right.
    \label{eq:P}
    \end{equation}
    Where $\mathcal{G}$ is the set of generator nodes, $\mathcal{R}$ is the set of renewable generation unit nodes, $\mathcal{T}_H$ is the set of high-voltage side nodes of the transformer, $\mathcal{T}_L$ is the set of low-voltage side nodes of the transformer, $\mathcal{N}(i)$ is the set of neighboring buses of bus $i$, $b(i,j)$ is the line branch or transformer branch terminated at buses $i$ and $j$, and $z_{b(i,j)}$ is the impedance of the branch.
    
    \item{\textbf{Topology Matrix $\mathbf{T}$:} The topology matrix represents the information about the connection relationships of the buses in the grid, as follows}
    \begin{equation}
    \mathbf{T}_{ij}=\left\{
    \begin{array}{ll}
    1 &,i=j~or~j \in \mathcal{N}(i) \\
    0 &,j \notin \mathcal{N}(i) \\
    \end{array}
    \right.
    \label{eq:T}
    \end{equation}
    
    \item{\textbf{Electrical Distance Matrix $\mathbf{D}_Z$:} The Electrical Distance Matrix represents the information about the electrical distance between the individual buses of the grid as follows, where $\mathbf{Z}$ denotes the node impedance matrix.}
    \begin{equation}
    {\mathbf{D}_Z}_{ij}=\left\{
    \begin{array}{ll}
    0 &,i=j \\
    \mathbf{Z}_{ij} &,i\ne j \\
    \end{array}
    \right.
    \label{eq:DZ}
    \end{equation}
    
    \item{\textbf{Graph Distance Matrix $\mathbf{D}$:} The Graph Distance Matrix represents the shortest distance information of each node on the graph under the graph structure, and the Dijkstra algorithm is a commonly used method to solve the shortest distance between two nodes. The Graph Distance Matrix is as follows.}
    \begin{equation}
    \mathbf{D}_{ij}=\left\{
    \begin{array}{ll}
    0 &,i=j \\
    \mathrm{Dijkstra}(i,j) &,i\ne j \\
    \end{array}
    \right.
    \label{eq:D}
    \end{equation}
\end{enumerate}

\subsection{Architecture of PGNN}
\begin{figure*}[!t]
\centering
\includegraphics[width=0.9\textwidth]{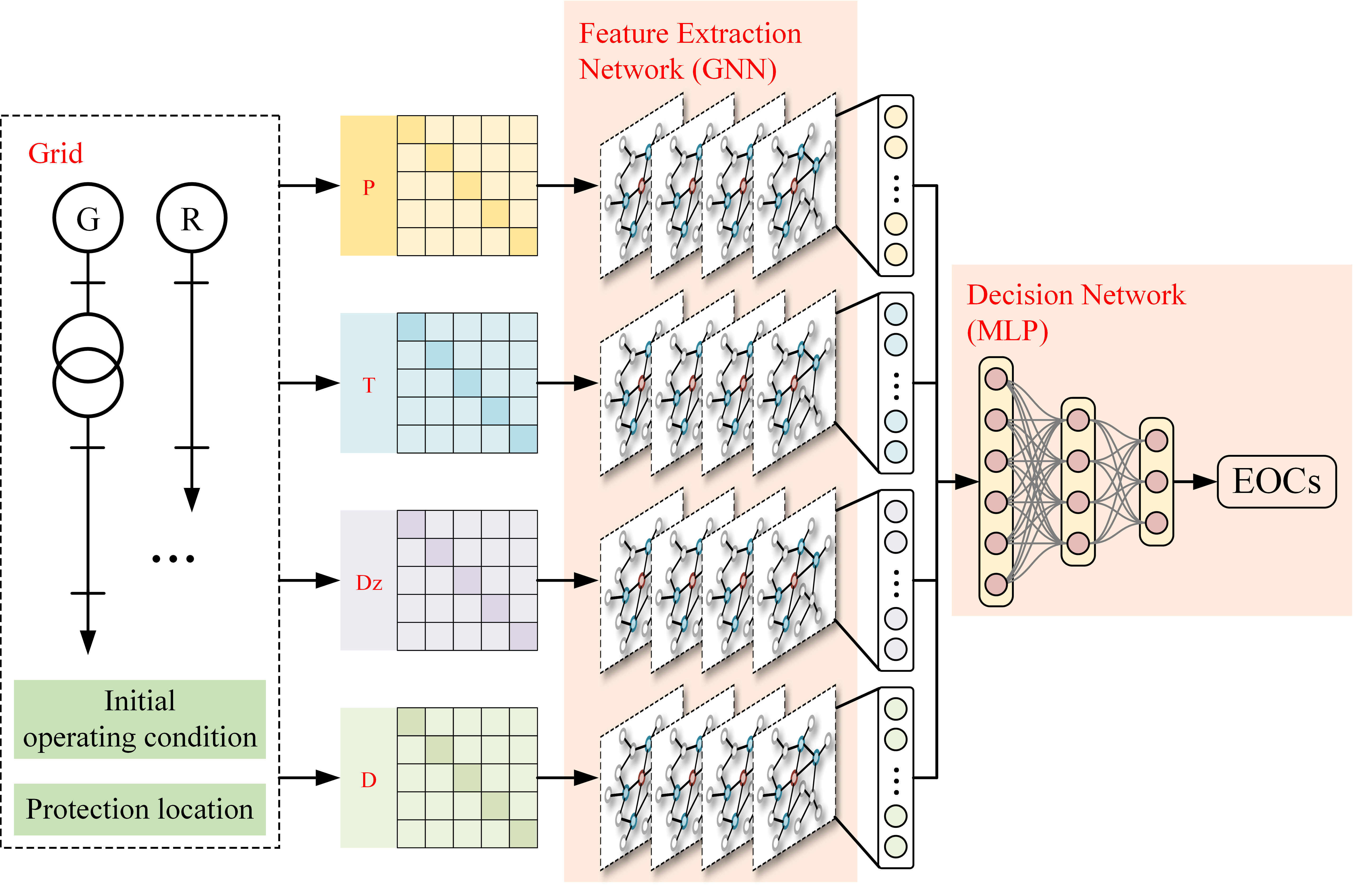}
\caption{Architecture of the PGNN.}
\label{fig:PGNN}
\end{figure*}

After extracting the power system information based on the parallel feature coding method, the PGNN framework is constructed in this section. The inputs of PGNN are the four matrices $\mathbf{P}$, $\mathbf{T}$, $\mathbf{D}_Z$, and $\mathbf{D}$ of the system, and the four matrices are input into four GNNs with the same structure. The output is the EOC of the current initial operating condition of the system. As shown in Fig. \ref{fig:PGNN}, the PGNN framework contains two parts: feature extraction network and decision network. The feature extraction network consists of four independent GNNs. Each of GNN contains several GraphSAGE layers, which are used to extract the four types of features of the system, respectively. The number of nodes of the GNN, as well as the dimensions of the input features of each node, are the same as the number of buses in the system. The outputs of feature extraction network are the extracted high-dimensional features of the power system.

The extracted high-dimensional features are spliced and stretched and then fed into the decision network, which is a Multi-Layer Perceptron (MLP). Assuming that the feature dimension of the node output from the feature extraction network is $d$, the number of buses in the system is $n$, and the number of lines is $m$. Considering the EOCs in the $N-k$ range of the system. $k$ is a predefined value, indicating that the number of transmission lines that can be out of service in the EOCs compared to the initial operating condition is at most $k$. The value of $k$ is usually taken as 1 to 3.
Then the input dimension of the decision network is $4dn$, and the output dimension is $m$, i.e., each output dimension corresponds to the on/off state of a line, with 1 being out of service and 0 being in service. 
The decision network is finally output through a Sigmoid function, which aims to ensure that all output values are positive. Taking the top $k$ values in the output greater than 0.5 corresponding to lines out of service, these $k$ lines are considered to be the lines that need to be out of service for the EOC. If there are less than $k$ values in the output that are greater than 0.5, take all of them corresponding to line outages.

A binary cross-entropy function is used as the loss function for supervised training of the PGNN network. Test and training data are obtained by enumeration. A number of randomly selected initial operating states of the system, including the on/off state $\bm{\tau}_0$ of lines and the location $p_l$ of the protection, are used as inputs to the PGNN. The labels are the EOCs that make the fault current flowing through the protection the largest, solved by brute-force enumeration. Set the lines that need to be shut down in the labels to 1 and the others to 0. The pseudo-code for the PGNN is shown in Algorithm \ref{alg:PGNN}.

The key enhancement of PGNN stems from its implementation of distinct parameter matrices ($\mathbf{W}_P$, $\mathbf{W}_T$, $\mathbf{W}_{D_Z}$, and $\mathbf{W}_D$) specifically designed for different input sources. This architectural design enables more comprehensive learning of fine-grained feature correlations within each matrix domain, while simultaneously mitigating potential interference caused by magnitude discrepancies among the four feature types.

\begin{algorithm}
    \caption{The Parallel Graph Neural Network Framework}
    \label{alg:PGNN}
    \textbf{Input :} Parallel Feature Matrices $\mathbf{P}$, $\mathbf{T}$, $\mathbf{D}_Z$, $\mathbf{D}$
    
    \textbf{Output:} $\bm{EOC}=[x_1',...,x_m']$

    \textbf{Symbol Description:} $N$ buses, $M$ transmission lines, $L$ layers of GraphSAGE for each GNN and consider EOCs in the $N-k$ range. $\mathbf{F}$ is the high-dimensional feature, and $\phi _D$ is the decision network.
    \BlankLine
    \BlankLine
    \textbf{Feature Extraction Network:}
    
    \For{\( l = 1 \) \KwTo \( L \)}{
        \For{\( node~n = 1 \) \KwTo \( N \)}{
        $\bm{P}_n^{(l+1)} = \sigma\left( \mathbf{W}_P^{(l)} \cdot \mathrm{AGG}^{(l)}\left( \{ \bm{P}_u^{(l)}, \forall u \in \mathcal{N}(n) \} \right) \right)$
        \BlankLine
        
        $\bm{T}_n^{(l+1)} = \sigma\left( \mathbf{W}_T^{(l)} \cdot \mathrm{AGG}^{(l)}\left( \{ \bm{T}_u^{(l)}, \forall u \in \mathcal{N}(n) \} \right) \right)$
        \BlankLine
        
        ${\bm{D}_Z}_n^{(l+1)} = \sigma\left( \mathbf{W}_{D_Z}^{(l)} \cdot \mathrm{AGG}^{(l)}\left( \{ {\bm{D}_Z}_u^{(l+1)}, \forall u \in \mathcal{N}(n) \} \right) \right)$
        \BlankLine
        
        $\bm{D}_n^{(l+1)} = \sigma\left( \mathbf{W}_D^{(l)} \cdot \mathrm{AGG}^{(l)}\left( \{ \bm{D}_u^{(l)}, \forall u \in \mathcal{N}(n) \} \right) \right)$
        }
    }

    \BlankLine
    \BlankLine
    \textbf{Decision Network:}

    $\mathbf{F}_{1\times 4dn}=\mathrm{Reshape}\left(
    \left[\mathbf{P}_{d\times d}, \mathbf{T}_{d\times d}, {\mathbf{D}_Z}_{d\times d}, \mathbf{D}_{d\times d} \right],-1 \right)$
    $=[\bm{P}_{1,:},\dots,\bm{P}_{d,:}, \bm{T}_{1,:},\dots,\bm{T}_{d,:},$
    $~~~~{\bm{D}_Z}_{1,:},\dots,{\bm{D}_Z}_{d,:}, \bm{D}_{1,:},\dots,\bm{D}_{d,:}   ]$

    $\bm{V}_{1\times M}=\mathrm{Sigmoid}\left(\phi _D\left(\bm{F} \right)\right)=\left[v_1,\dots,v_M\right]$

    \BlankLine
    \BlankLine
    \textbf{Calculate }$\bm{EOC}s$ \textbf{:}

    $\bm{V}' = \mathrm{Sort}(\bm{V}, \text{descending})=\left[v_1',\dots,v_M'\right] \Rightarrow v_1'\ge ,\dots,\ge v_M'$

    \For{\( i = 1 \) \KwTo \( M \)}{
      \uIf{\( v_i \geq v_k' \) \textbf{and} \( v_i \geq 0.5 \)}{
        \( EOC_i=x_i' \leftarrow 1 \)
      }
      \uElse{
        \( EOC_i=x_i' \leftarrow 0 \)
      }
    }
    \BlankLine

    \Return $\bm{EOC}_{1\times M}$
\end{algorithm}

\section{Experiments and Discussions}
\label{sec:EXPERIMENTS}
In this section, the proposed PGNN is validated in a selectvitity EOCs scenario with instantaneous overcurrent protection. The validity experiments and comparison experiments are performed on the IEEE 39-bus and 118-bus systems, respectively.

\subsection{Experimental Setup}
\label{sub:Setup}

\subsubsection{Experiment Environment}
Experiments are performed on a personal computer with a Intel(R) Core(TM) i7-10700F CPU @ 2.90GHz and 16 GB RAM. The proposed PGNN framework is implemented based on Python 3.10, Pytorch 2.0.1 and Pytorch Geometric 2.4.0.

\subsubsection{Performance Metrics}
To test the accuracy of the model in all aspects, this paper proposes 3 test metrics as follows

\begin{enumerate}
    \item{\textbf{Operating Conditions Accuracy (OC-Acc):} The \textbf{OC-Acc} is defined as the percentage of matching EOCs between PGNN outputs and brute-force enumeration results across $N_{test}$ randomly sampled initial states.}
    
    \item{\textbf{Short-Circuit Current Accuracy (SCC-Acc):} The \textbf{SCC-Acc} measures the percentage of matching short-circuit currents of EOCs between PGNN outputs and brute-force enumeration results across $N_{test}$ randomly sampled initial states. 
    To account for allowable calculation tolerance, the \textbf{e-SCC-Acc} is defined with parameter $e$ denoting the permissible error range (0.1 and 0.5 in this paper), where PGNN results within $\pm e$ of enumeration values are considered correct.}
    
    \item{\textbf{Protection Selectivity Accuracy (PS-Acc):} The \textbf{PS-Acc} measures the percentage of PGNN-calculated short-circuit currents of EOCs satisfying the protection selectivity criterion in \eqref{eq:PS-Acc}, evaluated across $N_{test}$ randomly sampled initial states. Where $K$ is taken as 1.2, $I_f^{PGNN}$ and $I_{f,max}^{Enum}$ is the short-circuit current calculated by PGNN and enumeration method, respectively. Failure to satisfy \eqref{eq:PS-Acc} results in $I_{set}$ encroaching downstream protection zones, violating protection selectivity constraints.}
    \begin{equation}
        I_{set}=KI_f^{PGNN}>I_{f,max}^{Enum}
    \label{eq:PS-Acc}
    \end{equation}
\end{enumerate}
The three performance metrics are in decreasing order of stringency, and the \textbf{PS-Acc} of the trained neural networks should reach 100\% to fully satisfy the practical requirements.

\subsection{Effectiveness Experiment}
\label{sub:Effectiveness Experiment}
In this section, experiments are conducted on the IEEE 39-bus and IEEE 118-bus systems, respectively, to verify the effectiveness of the proposed PGNN method. The IEEE 39-bus system contains 39 buses and 34 transmission lines. After combining the parallel double-circuit lines, the IEEE 118-bus system includes 118 buses and 169 transmission lines. Replace 5 (50\%) of the synchronous generators in the IEEE 39-bus system and 18 (33\%) in the IEEE 118-bus system with photovoltaic power plants.


The input dimension of the first GraphSAGE is the same as the number of buses in the power system, 39 and 118, respectively. The decision network $\phi _D$ is a three-layer and four-layer MLP, respectively, and the output dimension is the same as the number of transmission lines in the power system, which are 34 and 169. Except for the last layer of the MLP, which is output by a Sigmoid function, the rest of the GraphSAGE layers and the linear layers are followed by a ReLU activation function.

The hyperparameters, including the learning rate, optimizer, epoch, batch, training set size, and test set size are listed in Table \ref{tab:Training Parameters}. The samples are generated by the enumeration method, and each sample contains the initial operating state of the system and the EOC that makes the short-circuit current at the end of the line the largest. The initial operating states are randomly selected. The training set and the test set do not overlap to avoid overfitting and ensure the generalization ability of the neural network. The initial state of the system is set with the possibility of 0 to 2 lines out of service in this experiment, and the protection location is randomly selected among the service lines. The EOCs are in the range of $N-2$, i.e., the experiments in this section consider up to 4 out-of-service lines in the system.

\begin{table}[!t]
\caption{Training Parameters of PGNN\label{tab:Training Parameters}}
\centering
\setlength{\tabcolsep}{4pt}
\begin{tabular}{ccccccc}
\toprule
System & \makecell{Learning \\ rate} & Optimizer & Epoch & Batch & Train Set & Test Set\\
\midrule
39 & 0.001 & Adam & 10000 & 64 & 10000 & 2000\\
118 & 0.0005 & Adam & 20000 & 128 & 20000 & 4000\\
\bottomrule
\end{tabular}
\end{table}

\begin{figure}[!t]
\centering
\includegraphics[width=\columnwidth]{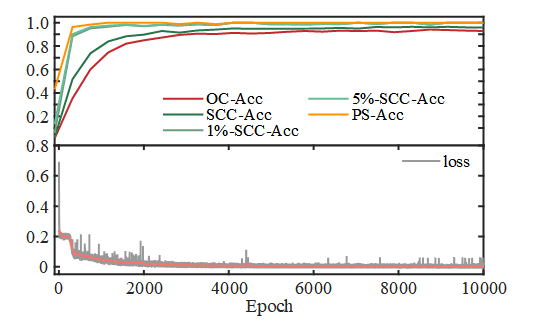}
\caption{Accuracy and Loss curves of PGNN on IEEE 39-bus System.}
\label{fig:acc-loss-39}
\end{figure}

\begin{figure}[!t]
\centering
\includegraphics[width=\columnwidth]{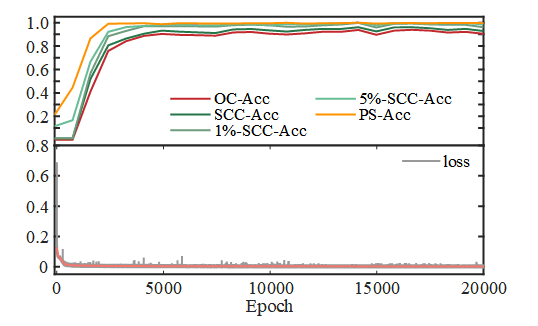}
\caption{Accuracy and Loss curves of PGNN on IEEE 118-bus System.}
\label{fig:acc-loss-118}
\end{figure}

The loss curves and accuracy curves of PGNN on both systems are shown in Fig. \ref{fig:acc-loss-39}, Fig. \ref{fig:acc-loss-118}, and Table \ref{tab:Performance}. The OC-Acc of PGNN is able to exceed 94\% on both systems and can ensure the PS-Acc to reach 100\% level, which proves the effectiveness of the method.

\begin{table}[!t]
\caption{Performance of PGNN on IEEE 39-bus And 118-bus Systems\label{tab:Performance}}
\centering
\setlength{\tabcolsep}{4pt}
\begin{tabular}{ccccccc}
\toprule
System & OC-Acc & SCC-Acc & 1\%-SCC-Acc & 5\%-SCC-Acc & PS-Acc\\
\midrule
39 & 94.1\% & 96.5\% & 99.5\% & 99.6\% & 100\%\\
118 & 94.6\% & 96.8\% & 98.9\% & 99.1\% & 100\%\\
\bottomrule
\end{tabular}
\end{table}

\subsection{Ablation Experiment}
\label{sub:Ablation}
The proposed PGNN method has two main improvements: replacing traditional CNNs with GNNs, which are more suitable for power system problems, and adopting a parallel feature coding approach and a parallel graph neural network structure. In this section, ablation experiments are conducted to remove the two improvement methods separately and verify their effectiveness on the IEEE 39-bus system. The algorithm is degraded to GNN, PCNN, and CNN after removal.

\begin{figure}[!t]
\centering
\includegraphics[width=\columnwidth]{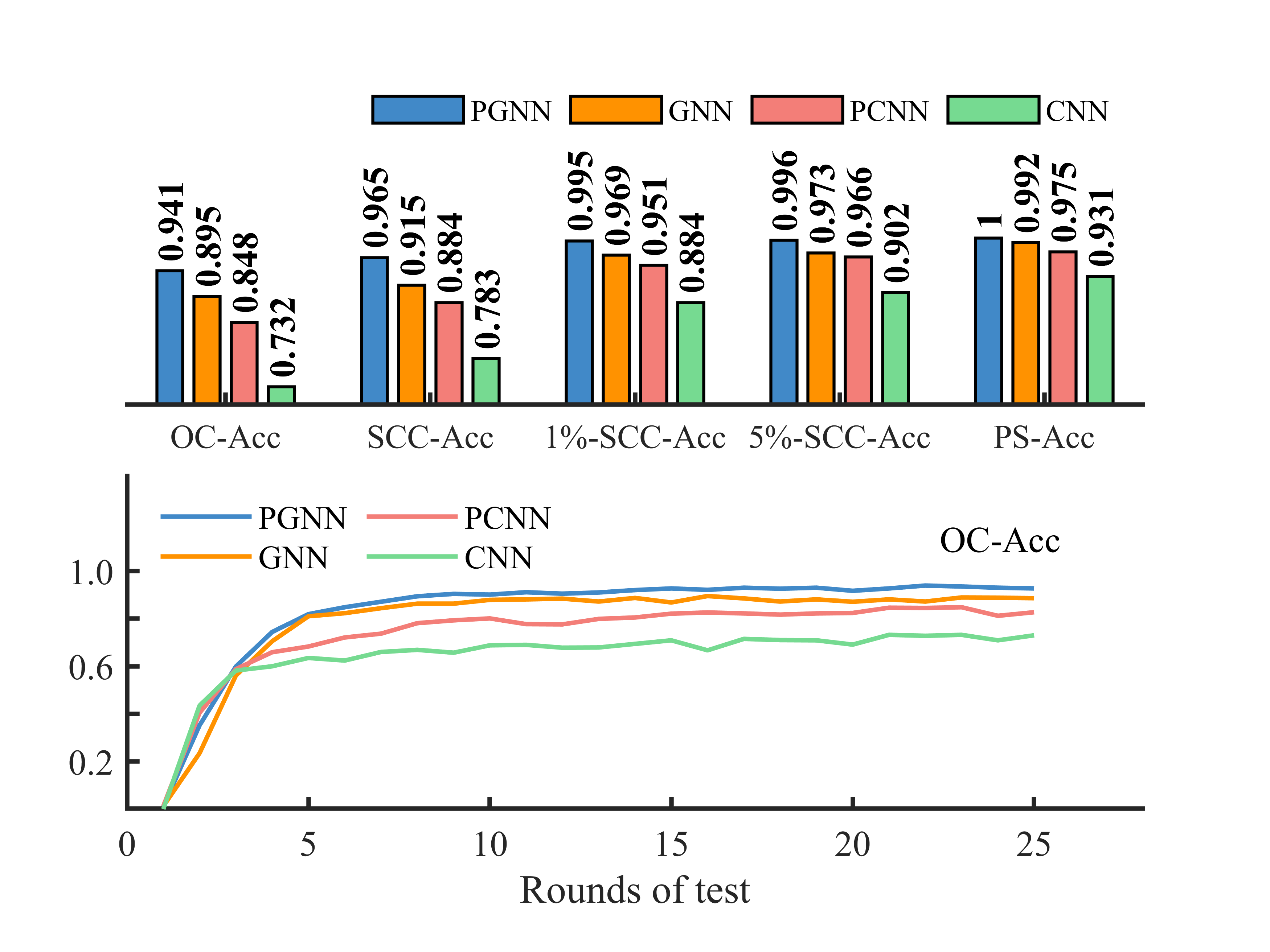}
\caption{Accuracy results of the ablation experiment.}
\label{fig:xiaorong}
\end{figure}

The accuracy rates are shown in Fig. \ref{fig:xiaorong}. The results show that the accuracy curves of the GNN, PCNN, and CNN methods are lower than PGNN, and the accuracy rates are PGNN, GNN, PCNN, and CNN in descending order. The results demonstrate that GNNs are more suitable for dealing with EOCS problems and that the parallel feature coding method and the parallel neural network structure can better extract the features of the power systems, which proves the effectiveness of the proposed improvements.

\subsection{Comparison Experiment}
\label{sub:Comparison}
In this section, comparative experiments are conducted to compare the performance of the proposed PGNN with existing EOCS methods in terms of different accuracy rates and inference time on IEEE 39-bus and 118-bus systems. Existing EOCS methods include enumeration methods (global enumeration method, local enumeration method) and heuristic algorithms (genetic algorithm as an example). The experiments also compared two neural network approaches, Parallel Graph Convolutional Networks (PGCN) and Parallel Graph Attention Networks (PGAT). The local enumeration method considers the operating conditions within a range of several levels of protection proximity, and we take a range of 3 levels in this paper. The Genetic Algorithm (GA) encodes the operating conditions as genetic genes, selects the genes that make the short-circuit current larger for inheritance and evolution, and finally obtains the EOCs. PGCN and PGAT are replacing GraphSAGEs with graph convolutional networks and graph attention networks, respectively.

The experimental results are shown in Table \ref{tab:Comparison}. In terms of accuracy, the accuracy of PGNN is higher than that of local enumeration, GA, PGCN and PGAT. 
Compared to local enumeration and global enumeration, the inference time of PGNN is reduced by 37 and 73 times, and 512 and 1697 times on IEEE 39-bus and 118-bus systems, respectively. 
The prolonged inference times coupled with suboptimal accuracy observed in genetic algorithm implementations for EOCS resolution demonstrate fundamental limitations in addressing this class of combination optimization problems.
The accuracy of PGNN is also higher than that of PGCN and PGAT. PGNN and PGCN demonstrate nearly identical inference times. PGAT exhibits the longest inference time among the three due to the attention mechanism.
Comparative evaluations confirm that the proposed PGNN framework exhibits superior accuracy rates and computational efficiency compared to conventional methods in EOCS problems.

\begin{table}[!t]
\caption{Comparison of Different Methods\label{tab:Comparison}}
\centering
\begin{tabular}{ccccccc}
\toprule
System & Method & OC-Acc & SCC-Acc & PS-Acc  & \makecell{Inference \\ Time}  \\
\midrule
\multirow{6}{*}{39} & PGNN & 94.1\% & 95.9\% & 100\% & 7.34ms\\
 & PGCN & 92.5\% & 93.2\% & 99.1\% & 7.38ms \\
 & PGAT & 91.3\% & 93.7\% & 99.2\% & 14.71ms \\
 & GA & 61.5\% & 63.6\% & 80.7\% & 3min2s \\
 & Local Enum & 88.5\% & 91.3\% & 100\% & 273.74ms \\
 & Global Enum & 100\% & 100\% & 100\% & 3.76s \\

\midrule
\multirow{6}{*}{118} & PGNN & 94.6\% & 96.8\% & 100\% & 25.86ms\\
 & PGCN & 92.5\% & 94.8\% & 99.5\% & 24.22ms \\
 & PGAT & 90.9\% & 92.2\% & 99.2\% & 52.32ms \\
 & GA & 55.8\% & 60.1\% & 72.3\% & 10min18s \\
 & Local Enum & 85.4\% & 88.2\% & 95.7\% & 1.88s \\
 & Global Enum & 100\% & 100\% & 100\% & 43.9s \\
\bottomrule
\end{tabular}
\end{table}

\section{Conclusion}
\label{sec:CONCLUSION}
This paper constructs a parallel feature encoding method and a parallel graph neural network to solve the Extreme Operating Condition Search (EOCS) problem in the online relay setting calculation of renewable power systems. The method encodes power system information from four perspectives: component parameters, network topology, electrical distance, and graph distance, and proposes a novel parallel graph neural network (PGNN) architecture to extract features. The proposed method is verified on the IEEE 39-bus and 118-bus systems. Experiments show that the proposed PGNN and the improvements are effective. Compared with the existing EOCS method, the inference time of the PGNN is reduced by 37 to 1697 times, while the accuracy rates exceed 94\%. 

The PGNN method proposed can adapt to changes in the operating conditions of several transmission lines in the system and has already demonstrated a certain level of generalization ability. Future investigations will explore the scalability of the proposed methodology under more significant structural changes in power systems, such as the integration of new components (e.g., lines, buses, and generators), while also addressing EOCS challenges associated with diverse protection principles.

\bibliographystyle{IEEEtran}
\bibliography{IEEEabrv,bibi}

\end{document}